\documentclass[conference]{IEEEtran}

\usepackage{cite}
\usepackage[numbers,sort&compress]{natbib}

\usepackage[utf8]{inputenc}
\usepackage{amsmath,amssymb,amsfonts}
\usepackage{graphicx}
\usepackage{bm}
\usepackage{enumerate}
\usepackage{comment}
\usepackage{xcolor}
\usepackage{url}
\usepackage{color,soul}

\usepackage{float}

\usepackage{algorithmic}
\usepackage{textcomp}

\usepackage{subfig}
\usepackage{booktabs}

\let\labelindent\relax
\usepackage[shortlabels]{enumitem}

\definecolor{light-gray}{gray}{0.8}

\def\BibTeX{{\rm B\kern-.05em{\sc i\kern-.025em b}\kern-.08em
    T\kern-.1667em\lower.7ex\hbox{E}\kern-.125emX}}

\makeatletter
\newcommand{\linebreakand}{%
  \end{@IEEEauthorhalign}
  \hfill\mbox{}\par
  \mbox{}\hfill\begin{@IEEEauthorhalign}
}

\newcommand{\rnum}[1]{\uppercase\expandafter{\romannumeral #1\relax}}

\begin{document}
\title{Contextual Hourglass Network for Semantic Segmentation of High Resolution Aerial Imagery \\
}

\author{
\small 

\begin{tabular}[t]{c@{\extracolsep{8em}}c} 

1\textsuperscript{st} Panfeng Li\footnotemark & 2\textsuperscript{nd} Youzuo Lin \\
\textit{Department of Electrical and Computer Engineering} & \textit{Center for Space and Earth Science} \\
\textit{University of Michigan} & \textit{Los Alamos National Laboratory} \\
Ann Arbor, USA & Los Alamos, USA \\
pfli@umich.edu & ylin@lanl.gov \\

\\

\multicolumn{2}{c}{3\textsuperscript{th} Emily Schultz-Fellenz}  \\
\multicolumn{2}{c}{\textit{Center for Space and Earth Science}} \\
\multicolumn{2}{c}{\textit{Los Alamos National Laboratory}}\\
\multicolumn{2}{c}{Los Alamos, USA} \\
\multicolumn{2}{c}{eschultz@lanl.gov} \\

\end{tabular}

}

\maketitle
\begin{abstract}

Semantic segmentation for aerial imagery is a challenging and important problem in remotely sensed imagery analysis. In recent years, with the success of deep learning, various convolutional neural network (CNN) based models have been developed. However, due to the varying sizes of the objects and imbalanced class labels, it can be challenging to obtain accurate pixel-wise semantic segmentation results. To address those challenges, we develop a novel semantic segmentation method and call it \textit{Contextual Hourglass Network}. In our method, in order to improve the robustness of the prediction, we design a new contextual hourglass module which incorporates attention mechanism on processed low-resolution featuremaps to exploit the contextual semantics. We further exploit the stacked encoder-decoder structure by connecting multiple contextual hourglass modules from end to end. This architecture can effectively extract rich multi-scale features and add more feedback loops for better learning contextual semantics through intermediate supervision. To demonstrate the efficacy of our semantic segmentation method, we test it on Potsdam and Vaihingen datasets. Through the comparisons to other baseline methods, our method yields the best results on overall performance.

\end{abstract}
\section{Introduction}
\label{sec:intro}

\begin{figure*}[ht]
    \centering
    \subfloat[a][Overview of Contextual Hourglass Network (CxtHGNet)]{
       \includegraphics[width=1\linewidth]{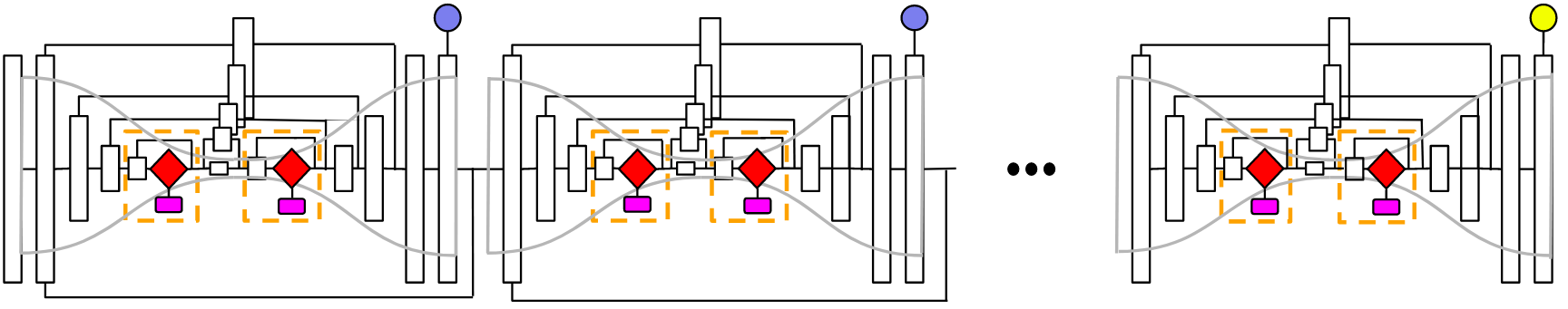}
       \label{fig:CxtHGNet}
    }

    \subfloat[b][An Illustration of Encoding Procedure. The variable of $C$ is the number of features, and $D$ is the number of categories in dataset.]{
       \includegraphics[width=1\linewidth]{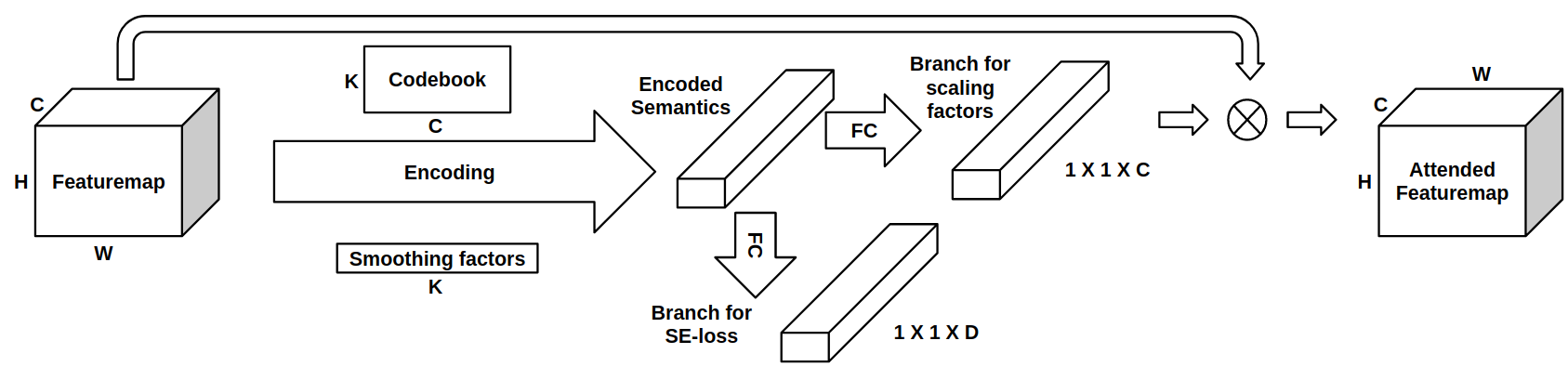}
       \label{fig:encoding}
    }
    \caption[]{(a) Each white box in the figure corresponds to a residual block~\cite{He_2016_CVPR}. Blue circles are the intermediate predictions whereas the yellow one is the final prediction. A loss function is applied to all these predictions through the same ground truth. The region in the dashed orange box represents the encoding procedure, where red rhombus is the context layer and the pink box is the branch for semantic encoding loss. (b) The Encoding Layer contains a codebook and smoothing factors, capturing encoded semantics. The top branch predicts scaling factors selectively highlighting class-dependent featuremaps. The down branch predicts the presence of the categories in the scene. (Notation: FC fully connected layer, $\otimes$ channel-wise multiplication.)}
    \label{fig:network}
\end{figure*}

Semantic segmentation becomes one of the most important problems in remotely sensed aerial imagery analysis. In particular, semantic segmentation can be applied to change detection, urban planning, or even automatic map mapping. Compared to the semantic segmentation on natural images, this task can be much more challenging for remotely sensed high resolution aerial imagery due to the high spatial resolution and large volumes of pixels. In particular, the size of the objects in the high-resolution imagery usually varies significantly. Due to the viewing perspectives, some objects in a certain area captured can be obscure and dim. Not to mention that many similarities may exist among different categories and large variances may happen to the same category. To achieve good performance on semantic segmentation of high resolution aerial imagery, the segmentation models should have the following two characteristics:

\begin{itemize}[itemsep=0em,topsep=0pt,parsep=0pt,partopsep=0pt,leftmargin=*,labelindent=5pt]
  \item Extraction of rich features across multi-scales to capture relatively small objects.
  \item Utilization of contextual semantics which is significant for distinguishing objects of varying sizes.
\end{itemize}

Recently, various deep neural networks structures have been utilized for semantic segmentation in aerial and medical imagery. In \cite{yao2024building, xiao2024convolutional, Kaiser_2017_TGRS, zhu2021pseudo, Liu_2017_CVPRW}, Fully Convolutional Networks (FCN)~\cite{Long_2015_CVPR} have been used as the backbone of their networks. Audebert \textit{et al.} \cite{audebert_beyond_2017} further utilizes SegNet~\cite{Badrinarayanan_2017_TPAMI} for the segmentation task which is an adaption of FCN by replacing the decoder with a series of pooling and convolutional layers. To overcome the loss of information from the initial layers and combine features from different scales, skip connection have been used in~\cite{Li_2017_ICUS, Mou2018VehicleIS} for high resolution imagery segmentation. However, the utilization of skip layers may not be enough for extracting rich multi-scale features. Recent work~\cite{Newell_2016_ECCV} shows that by stacking multiple encoder-decoders from end to end, enabling repeated bottom-up, top-down inference across various scales, the network performance is greatly improved. In this paper, we adopt the stacked encoder-decoder structure as the backbone of our network to extract rich multi-scale features.

Besides the pixel-level information, how to utilize contextual information is a key point for semantic labeling. Contextual relationships provide valuable information from neighborhood objects. Recently, channel-wise~\cite{Zhang_2018_CVPR}, point-wise~\cite{Zhao_2018_ECCV} attention mechanisms or their combination~\cite{Fu_2018_DANet} have been utilized for exploiting the contextual semantics in semantic segmentation and achieved state-of-art results. However, those approaches are based on FCN and they do not exploit the stacked encoder-decoder structure. Their performances are affected due to the lack of intermediate supervision on the predictions of multiple encoders, which has been demonstrated to be effective in prior work~\cite{Newell_2016_ECCV, Wei_2016_CVPR, Carreira_2016_CVPR, yufeng-24, yan2024survival, wang2024research}. 

In this work, in order to improve the overall performance, we develop a novel semantic segmentation method for high-resolution aerial imagery. The structure of our method is based on the original hourglass module~\cite{Newell_2016_ECCV} with several major differences. The main contributions of our work can be summarized as follows:

1) We design a novel contextual hourglass module which incorporates attention mechanism on processed low-resolution featuremaps to exploit the contextual semantics and therefore improve the robustness of the prediction. 

2) We exploit the stacked encoder-decoder structure by connecting multiple contextual hourglass modules from end to end. This architecture can effectively extract features from various scales and increase feedback loops for better learning contextual semantics through intermediate supervision.

Due to the connection to the original hourglass network, we call our network structure, \textit{Contextual Hourglass Network~(CxtHGNet)}. To validate the performance of our CxtHGNet, we test it on both Potsdam and Vaihingen open datasets acquired for the purposes of urban classification and semantic labeling. The test results show that our CxtHGNet can not only capture small-sized objects but also produce more consistent results.

\section{Approach}
\label{sec:approach}

In this section, we will provide the details of the contextual hourglass module and our Contextual Hourglass Network~(CxtHGNet).

\subsection{Contextual Hourglass Module}

Our contextual hourglass module is a symmetric structure inspired by hourglass module~\cite{Newell_2016_ECCV}. It firstly processes features down to a low resolution by a set of convolutional and pooling layers, then applies channel-wise, point-wise or other attention mechanisms on the processed low-resolution featuremap, and finally bi-linearly upsamples and combines features until reaching the final output resolution. The location we choose to apply the attention mechanism should make it retain more details without greatly increasing the computational cost. The contextual hourglass module is then boosted by its inner attention mechanism which utilizes contextual information to improve the labeling robustness.

In this work, we choose to utilize the encoding layer~\cite{Zhang_2018_CVPR, Zhang_2017_CVPR} as our channel-wise attention mechanism to test the network performance, which has the ability to selectively highlight class-dependent featuremaps. It may be worthwhile to mention that it is convenient to replace the channel-wise attention used here with other attention mechanisms~\cite{dai-23, ning2022rapid} which means our contextual hourglass module is generalizable.

The encoding procedure is shown in Figure~\ref{fig:encoding}. The encoding layers are placed at the locations where the featuremaps reach the $1/8$ size of the original input. The input of encoding layer are featuremaps within contextual hourglass module, of shape $H \times W \times C$, which corresponds to a set of C-dimensional input features $X = \{x_1, \dots, x_N\}$, where $N = H \times W$. The layer has a learnable inherent codebook $D = \{d_1, \dots, d_K\}$ containing $K$ number of codewords (visual centers) and a set of smoothing factor of the visual centers $S = \{s_1, \dots, s_K\}$. The output of encoding layer is the residual encoder $E = \{e_1, \dots, e_K\}$ of shape $K \times C$, and $e_k = \sum_{i=1}^N e_{ik}$, where $e_{ik}$ aggregates the residuals with soft-assignment weights, namely

\begin{equation}\label{eq:encoding}
    e_{ik} = \frac{\exp \left(-s_k \left\Vert r_{ik} \right\Vert^2 \right)}{\sum_{j=1}^K \exp \left(-s_j \left\Vert r_{ij} \right\Vert^2 \right)} r_{ik},
\end{equation}

\noindent where the residuals are given by $r_{ik} = x_i - d_k$. The final encoded semantics $e$ is summed up over $K$ residual encoders, namely $e = \sum_{k=1}^K \phi(e_k)$, where $\phi$ denotes ReLU activation.

As shown in Figure~\ref{fig:CxtHGNet}, there are two encoding layers within each contextual hourglass module, sharing the same codebook $D$ and smoothing factors $S$. Upon each encoding layer, two branches are further applied. One stacks a fully connected layer on it with a sigmoid activation function and outputs the scaling factors $\gamma$. The channel-wise multiplication between input $X$ and scaling factors $\gamma$, namely $Y = X \otimes \gamma$, emphasizes or de-emphasizes the class-dependent featuremaps. Another branch also stacks a fully connected layer with a sigmoid activation function on the encoding layer, which outputs individual predictions for the presences of object categories in an image and learns with a binary cross entropy loss, namely semantic encoding loss.

\subsection{Contextual Hourglass Network (CxtHGNet)}

The CxtHGNet connects multiple contextual hourglass modules end-to-end consecutively, as shown in Figure~\ref{fig:CxtHGNet}. It enables repeated bottom-up, top-down inference across various scales and consolidates global and local information of the whole image. In this paper, we stack four contextual hourglass modules in our network and the number of output features in each contextual hourglass module is 128, 128, 256, and 256 at corresponding locations where the resolution drops. The final output is then obtained by summing up the outputs of all contextual hourglass modules.

After the featuremaps being emphasized or de-emphasized by the inner attention mechanism, CxtHGNet will reuse the featuremaps for later stacking of contextual hourglass modules. As a result, CxtHGNet not only allows further exploiting higher order spatial relationships by processing the high-level features again but also adds more feedback loops for learning contextual semantics through intermediate supervision.

\begin{figure*}[ht]
	\centering
    \includegraphics[width=1\linewidth]{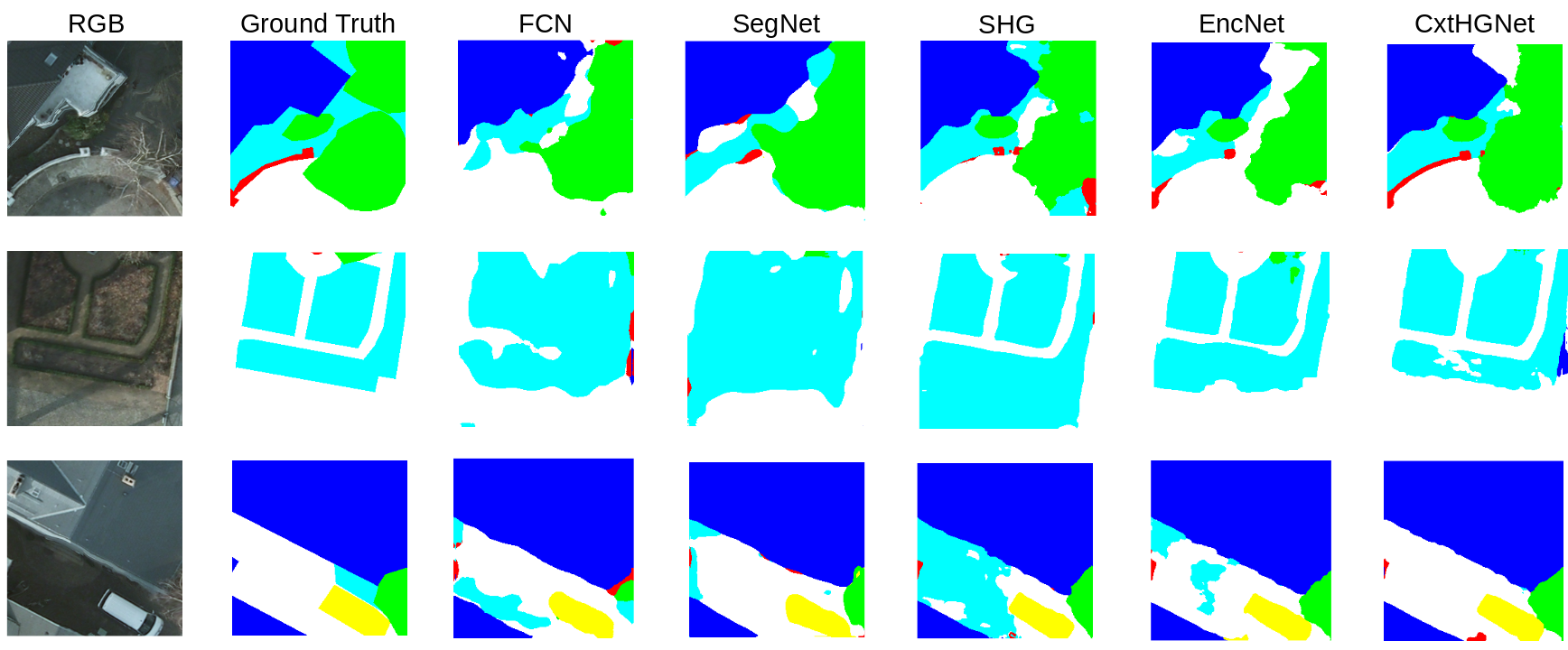}
    \caption{Selected results on Potsdam and Vaihingen test set.}
    \label{fig:results}
\end{figure*}

\section{Experimental Details}
\label{sec:results}

In this section, we first briefly describe the dataset, then provide the implementation details and show our final results.

\subsection{Data}

The Potsdam and Vaihingen 2D segmentation dataset are provided by Commission \rnum{2} of ISPRS~\cite{Rottensteiner_2012_ISPRS}. The Potsdam dataset includes 38 high-resolution aerial images, where 24 images are used for training. Each image has the size of 6,000 $\times$ 6,000 and contains 5 channels, namely near-infrared (NIR), red (R), green (G), blue (B), and the normalized digital surface models (nDSMs). The Vaihingen dataset consists of 33 high-resolution aerial images, 16 of which are given labels for training. Compared to Postdam dataset, the image data have different resolutions and do not have the Blue (B) channel.

We propose an online sliding window method to extract patches of size 256 $\times$ 256 from the original images without overlap, and pad 0s if needed. We further split the original training images to training and validation sets under the ratio of 9:1. Finally, in the Potsdam dataset, there are 12,441 images for training, 1,383 images for validation and 8,064 images for test, whereas in the Vaihingen dataset, there are 1,191 images for training, 133 for validation and 1,514 for test. 
 
\subsection{Implementation Details}

We implement our network based on open-source toolbox Tensorflow~\cite{tensorflow} and train it on 4 NVIDIA GTX 1080 Ti GPUs. Each GPU processes a batch of 4 images and the total batch size is 16. The training data are randomly shuffled and we drop the remainder of the last batch. We use the learning rate scheduling $lr = baselr * (1 - \frac{iter}{total\_iter})^{power}$ following prior work~\cite{Zhao_2017_CVPR, Zhang_2018_CVPR}. The Adam optimizer~\cite{Adam} is used for optimization, and we set the base learning rate as $0.0001$, the power as $0.95$. The image data in Potsdam and Vaihingen datasets have 5 and 4 channels where pretrained weights are unavailable. We follow~\cite{Zhang_2018_CVPR} and use a similar way, namely first pretrain the network without encoding layer for 100 epochs, then restore the pretrained weights and train our CxtHGNet another 100 epochs.

For data augmentation, we randomly flip the image horizontally and vertically, then scale it between 0.5 to 2 and finally crop the image into fix size padding 0s if needed.

\subsection{Results on Potsdam and Vaihingen Datasets}

We train our CxtHGNet on the training set and evaluate it on the test set using two standard metrics, namely pixAcc and mIoU. The validation set is used for adjusting hyperparameters in the network.

We use FCN~\cite{Long_2015_CVPR}, SegNet~\cite{Badrinarayanan_2017_TPAMI}, Stacked Hourglass Network (SHG)~\cite{Newell_2016_ECCV} and EncNet~\cite{Zhang_2018_CVPR} as our baseline models. FCN is the generally used framework for semantic segmentation, and SegNet is an adaptation of FCN by replacing the decoder with a series of pooling and convolution layers. SHG and EncNet are compared with our CxtHGNet to show the effectiveness of attention mechanism within contextual hourglass modules and the stacked encoder-decoder structure.

\begin{table}[h]
\centering
\resizebox{0.9\linewidth}{!}
  {
  \begin{tabular} {l c c | c }
    \toprule[1pt]
    {\bf Method} & {\bf Backbone} & {\bf pixAcc\%} & {\bf mIoU\%} \\
    \hline \hline
    FCN~\cite{Long_2015_CVPR} & VGG-16 & 82.75 & 61.71 \\
    SegNet~\cite{Badrinarayanan_2017_TPAMI} & VGG-16 & 83.93 & 63.42 \\
    SHG~\cite{Newell_2016_ECCV} & Hourglass-104 & 85.38 & 67.26 \\
    EncNet~\cite{Zhang_2018_CVPR} & ResNet-101 & 86.52 & 69.45 \\
    CxtHGNet (ours) & Hourglass-104 & \bf{87.15} & \bf{70.28} \\
    \bottomrule[1pt]
  \end{tabular}
  }
\caption{Numerical results on Potsdam test set.}
\vspace{-0.8em}
\label{tab:potsdam}
\end{table}

\begin{table}[h]
\centering
\resizebox{0.9\linewidth}{!}
  {
  \begin{tabular} {l c c | c }
    \toprule[1pt]
    {\bf Method} & {\bf Backbone} & {\bf pixAcc\%} & {\bf mIoU\%} \\
    \hline \hline
    FCN~\cite{Long_2015_CVPR} & VGG-16 & 82.39 & 61.58 \\
    SegNet~\cite{Badrinarayanan_2017_TPAMI} & VGG-16 & 83.64 & 63.23 \\
    SHG~\cite{Newell_2016_ECCV} & Hourglass-104 & 85.42 & 67.20 \\
    EncNet~\cite{Zhang_2018_CVPR} & ResNet-101 & 86.63 & 69.72 \\
    CxtHGNet (ours) & Hourglass-104 & \bf{87.26} & \bf{70.50} \\
    \bottomrule[1pt]
  \end{tabular}
  }
\caption{Numerical results on Vaihingen test set.}
\vspace{-0.8em}
\label{tab:vaihingen}
\end{table}

For the Potsdam dataset, there are about 12,000 training samples and 8,000 test samples, the training versus test ratio is about 1.5:1. The numerical results are shown in Table~\ref{tab:potsdam}. We can see that FCN and SegNet produce the worst results. SHG and EncNet yield considerably improved accuracies and our CxtHGNet achieves the best performance with $87.15\%$ pixAcc and $70.28\%$ mIoU and it outperforms all baseline models.

For the Vaihingen dataset, there are about 1,200 training samples and 1,500 test samples, the training versus test ratio is about 0.8:1. From Table~\ref{tab:vaihingen}, similar to the results on Potsdam dataset, FCN and SegNet both produce the worst results. SHG and EncNet improve the performance while out CxtHGNet produces the best results. Specifically, CxtHGNet achieves $87.26\%$ pixAcc and $70.50\%$ mIoU, showing our model also works well when the training data are limited.

We also show some visual examples of images with a size of $256 \times 256$ in Fig.~\ref{fig:results}. In the 1st and 2nd images, CxtHGNet captures the small red and white regions well, whereas FCN and SegNet are almost not capable of extracting these small-sized regions. In the 3rd image, we can see that compared to all other models, CxtHGNet predicts more consistent results. However, CxtHGNet also has its weaknesses, in the 1st image, CxtHGNet yields wrong prediction for the top middle area, and despite capturing well the white small region, CxtHGNet provides noisy predictions on the 2nd image.

\section{Conclusion}
\label{sec:Conclusion}
We develop a novel Contextual Hourglass Network (CxtHGNet) for semantic segmentation of high-resolution aerial imagery. Our CxtHGNet can extract rich multi-scale features of the image and learn the contextual semantics in scenes, due to the incorporation of bottom-up, top-down inference across various scales, attention mechanism, and intermediate supervision.  Also, It may be worthwhile to mention that with our CxtHGNet it can be convenient to replace the channel-wise attention used in this paper with other attention mechanisms and it makes our network structure more generalizable for other applications. The experimental results on Potsdam and Vaihingen datasets have demonstrated the superiority of our CxtHGNet.

\section*{Acknowledgment}
This work was funded by the Center for Space and Earth Science at Los Alamos National Laboratory.

\renewcommand{\bibfont}{\footnotesize}

\footnotesize{
\bibliographystyle{IEEEtran}
\bibliography{main}
}

\end{document}